\documentclass[runningheads]{llncs}
\usepackage{graphicx}
\usepackage[utf8]{inputenc} % allow utf-8 input
\usepackage[T1]{fontenc}    % use 8-bit T1 fonts
\usepackage{hyperref}       % hyperlinks
\usepackage{url}            % simple URL typesetting
\usepackage{booktabs}       % professional-quality tables
\usepackage{nicefrac}       % compact symbols for 1/2, etc.
\usepackage{microtype}      % microtypography
\usepackage{xcolor}         % colors
\usepackage{algorithm}
\usepackage[noend]{algpseudocode}
\usepackage{wrapfig}
\usepackage{amsmath,amsfonts,amssymb}
\usepackage{multirow}
\usepackage{subcaption}
\usepackage[title]{appendix}

\begin{document}
\title{Discrete Denoising Diffusion Approach to Integer Factorization\thanks{Supported by Latvian Council of Science, project “Smart Materials, Photonics, Technologies and Engineering Ecosystem”, project No. VPP-EM-FOTONIKA-2022/1-0001; Latvian Quantum Initiative under European Union Recovery and Resilience Facility project no. 2.3.1.1.i.0/1/22/I/CFLA/001; the Latvian Council of Science project lzp-2021/1-0479; Google Research Grants; NVIDIA Academic Grant "Deep Learning of Algorithms". }}
%
%\titlerunning{Abbreviated paper title}
% If the paper title is too long for the running head, you can set
% an abbreviated paper title here
%
\author{Kārlis Freivalds\inst{1} \and
Emīls Ozoliņš\inst{2}\and
Guntis Bārzdiņš\inst{2}}
\authorrunning{K.Freivalds et al.}
% First names are abbreviated in the running head.
% If there are more than two authors, 'et al.' is used.
%
\institute{Institute of Electronics and Computer Science, Latvia \\
\email{karlis.freivalds@edi.lv} \and
Institute of Mathematics and Computer Science \\
}
\maketitle              % typeset the header of the contribution

\begin{abstract}
  Integer factorization is a famous computational problem unknown whether being solvable in the polynomial time. With the rise of deep neural networks, it is interesting whether they can facilitate faster factorization. We present an approach to factorization utilizing deep neural networks and discrete denoising diffusion that works by iteratively correcting errors in a partially-correct solution. To this end, we develop a new seq2seq neural network architecture, employ relaxed categorical distribution and adapt the reverse diffusion process to cope better with inaccuracies in the denoising step. The approach is able to find factors for integers of up to 56 bits long. Our analysis indicates that investment in training leads to an exponential decrease of sampling steps required at inference to achieve a given success rate, thus counteracting an exponential run-time increase depending on the bit-length. 
\end{abstract}

\section{Introduction}
Deep Neural Networks have shown excellent results not only for real-world tasks but also for intellectually demanding algorithmic tasks such as sorting and multiplication \cite{freivalds2017improving}, NP-hard problems including Boolean Satisfiability (SAT)\cite{ozolins2022goal}, Travelling Salesman Problem (TSP) \cite{gaile2023unsupervised} and game playing \cite{silver2018general}. 
But there are algorithmic tasks that are too complex for a neural network to predict the solution directly. One of such tasks is integer factorization (the inverse of multiplication) where the goal is to find the prime factors of an integer number. Integer factorization \cite{lenstra2000integer} is a famous computational problem believed not to be solvable in polynomial time but also suspected that it is not NP-complete. There exists a fast quantum algorithm \cite{Shor1994AlgorithmsFQ} for factorization but it is questionable whether a sufficiently capable quantum computer can be built. Therefore it is tempting to find out whether neural networks can facilitate finding the factors quickly and this is the subject of our research. 

The best algorithm is General Number Field Sieve which, for factoring a $b$ bit number, is of complexity $\exp((64/9)^{1/3}+o(1))b^{1/3}\log(b)^{2/3})$ \cite{buhler1993factoring}. Simple algorithms can factor small integers quickly, for example, Pollard's rho algorithm \cite{pollard1978monte} factorizes a 56-bit number in about 0.03 seconds but struggles on large numbers. The largest factored cryptographic-hard number is 829 bits long. 

Some attempts have been made to predict integer factors using neural networks. In \cite{meletiou2002first} experiments were done with tiny neural networks and encoding the input and output as a single real number. In \cite{jansen2005neural} binary encoding was used reaching 37\% success at factoring 20-bit numbers using a neural network only and 90\% when followed by brute-force post-processing to correct up to 4 bits in the solution.

In this paper, we propose an indirect way of approaching integer factorization that requires the neural network to learn a simpler task i.e. to correct errors in a given partially-correct solution.
Then, at inference time such error correction is applied iteratively in a randomized fashion to search for a fully-correct solution. Our approach is derived from Denoising Diffusion \cite{sohl2015deep} which gives a strong theoretical basis for such a randomized search strategy. Diffusion models naturally allow sampling from the entire solution distribution, rather than only giving the most probable solution -- a feature needed for our task since there can be many ways to factor a given number. Also, diffusion models can be conditioned on subsidiary data -- the number to be factored, in our case.  

We adapt the diffusion algorithm so that it works well for factorization. First, we modify the sampling algorithm to retain the full probability information from step to step and sample only when presenting data to the denoising neural network. Second, we relax the discrete distribution using Gumbel-Softmax \cite{jang2016categorical} technique to make the denoising task easier to learn. We show that these modifications improve the factorization performance. For the denoising task, we evaluate several existing neural architectures and develop a new one that outperforms the existing ones.

We evaluate our approach on integers up to 56 bits long. The success of factoring a given number depends on the number of sampling steps that we perform at the inference and in this paper we give detailed analysis with respect to different number of sampling steps and bit lengths. For example, we get 98\% correctly factored 32-bit numbers in 8192 sampling steps and 31\% correctly factored 40-bit numbers given 16384 sampling steps. We also evaluate the scaling behavior which reveals two trends: (a) longer numbers require exponentially more sampling steps and (b) longer training results in an exponential decrease of required sampling steps.  

% Our main contributions, in summary:
% \begin{itemize}
%     \item Adapting the Denoising Diffusion approach for solving the Integer Factorization problem;
%     \item Introducing a new, powerful neural model for seq2seq tasks;
%     \item Using relaxed categorical distribution for diffusion;
%     \item Estimating the scaling behavior with respect to training time, sampling time, and bit-length. 
% \end{itemize}

\section{Background: Diffusion Models}
\label{sec:background}
Diffusion models have achieved state-of-the-art results for image generation \cite{ho2022cascaded,vahdat2021score,rombach2021high}. Diffusion has been applied to discrete binary \cite{sohl2015deep} and categorical \cite{hoogeboom2021argmax} data, text generation \cite{austin2021structured}. But not for such hard combinatorial problems like integer factorization. 

Given data $x_0$, a diffusion model \cite{sohl2015deep} consists of predefined variational distributions $q(x_t|x_{t-1})$ that gradually introduces noise over time steps $t \in \{1,...,T\}$. The diffusion trajectory is defined such that $q(x_t|x_{t-1})$ adds a small amount of noise around $x_t$.
This way, information is gradually destroyed and at the final time step, $x_T$ carries almost no information about $x_0$. A nice property of the diffusion process is that it can be reversed if the gradient of the distribution can be estimated which is often expressed as a function that denoises the data. Usually, Normal distribution is employed which is simple to work with and produces excellent results for images \cite{ho2022cascaded} and sound \cite{Diffwave}.

To deal with discrete values, here we employ diffusion for categorical data, namely the Multinomial Diffusion \cite{hoogeboom2021argmax}. Having $K$ categories, $x_t$ is encoded as one-hot vector $x_t \in\{0,1\}^K$. The multinomial diffusion process is defined using a categorical distribution that has a small probability $\beta_t$ of resampling a category uniformly and a large $(1-\beta_t)$ probability of sampling the previous value $x_{t-1}$:

\begin{equation}
q(x_t|x_{t-1}) = \mathcal{C}(x_t|(1-\beta_t)x_{t-1}+\beta_t/K),
\end{equation}

where $\mathcal{C}$ denotes a categorical distribution with probability parameters after $|$. For such diffusion process the probability of any $x_t$ given $x_0$ is expressed as: 
\begin{equation}
q(x_t|x_0) = \mathcal{C}(x_t|\bar{\alpha}_t x_0+(1-\bar{\alpha}_t)/K),
\label{eq:qsample}
\end{equation}
where $\alpha_t=1-\beta_t$ and $\bar{\alpha}_t=\prod_{\tau=1}^t{\alpha_\tau}$. For reverse distribution step, we follow the common practice to parametrize it using $x_0$.  According to \cite{hoogeboom2021argmax}, the distribution for the previous time step $t-1$ can be computed from the value $x_t$ at the next step and the initial value $x_0$ as:

\begin{gather}
q(x_{t-1}|x_t, x_0) = \mathcal{C}(x_{t-1}|\theta_{post}(x_t, x_0)),\\
\theta_{post}(x_t, x_0)=\tilde{\theta}/\sum_{k=1}^K{\tilde{\theta}_k},\\
\tilde{\theta}=[\alpha_t x_t + (1-\alpha_t)/K]\odot[\bar{\alpha}_{t-1} x_0 + (1-\bar{\alpha}_{t-1})/K]
\label{eq:3}
\end{gather}

% \begin{equation}
% \begin{split}
% q(x_{t-1}|x_t, x_0) = \mathcal{C}(x_{t-1}|\theta_{post}(x_t, x_0)), \\
% \text{where } \theta_{post}(x_t, x_0)=\tilde{\theta}/\sum_{k=1}^K{\tilde{\theta}_k},\\
% \text{and } \tilde{\theta}=[\alpha_t x_t + (1-\alpha_t)/K]\odot[\bar{\alpha}_{t-1} x_0 + (1-\bar{\alpha}_{t-1})/K]
% \end{split}
% \label{eq:3}
% \end{equation}
During the reverse process, an approximation $\hat{x}_0$ is used instead of $x_0$ which is produced by a neural network $\mu$: $\hat{x}_0=\mu(x_t, \bar{\alpha}_t)$. The neural network is trained by feeding it with the cumulative noise\footnote{\cite{hoogeboom2021argmax} parametrize the neural network with $t$, instead. This is equivalent once we fix the noise schedule.} at time $t$ and noisy sample $x_t$ which is produced by the forward diffusion and asking the network to produce a clean sample $\hat{x}_0$. We use a linear schedule of $\bar{\alpha}$ both during training and inference. The loss function for training is the KL divergence between the true distribution and the predicted one: 
\begin{equation}
KL(\mathcal{C}(\theta_{post}(x_t, x_0))|\mathcal{C}(\theta_{post}(x_t, \hat{x}_0)))
\label{eq:KL}
\end{equation}

\section{Diffusion for Factorization}\label{sec:factoring}
\begin{figure*}[!ht]
    \centering
    \begin{minipage}[t]{.48\textwidth}
        \centering
        \includegraphics[width=1\columnwidth]{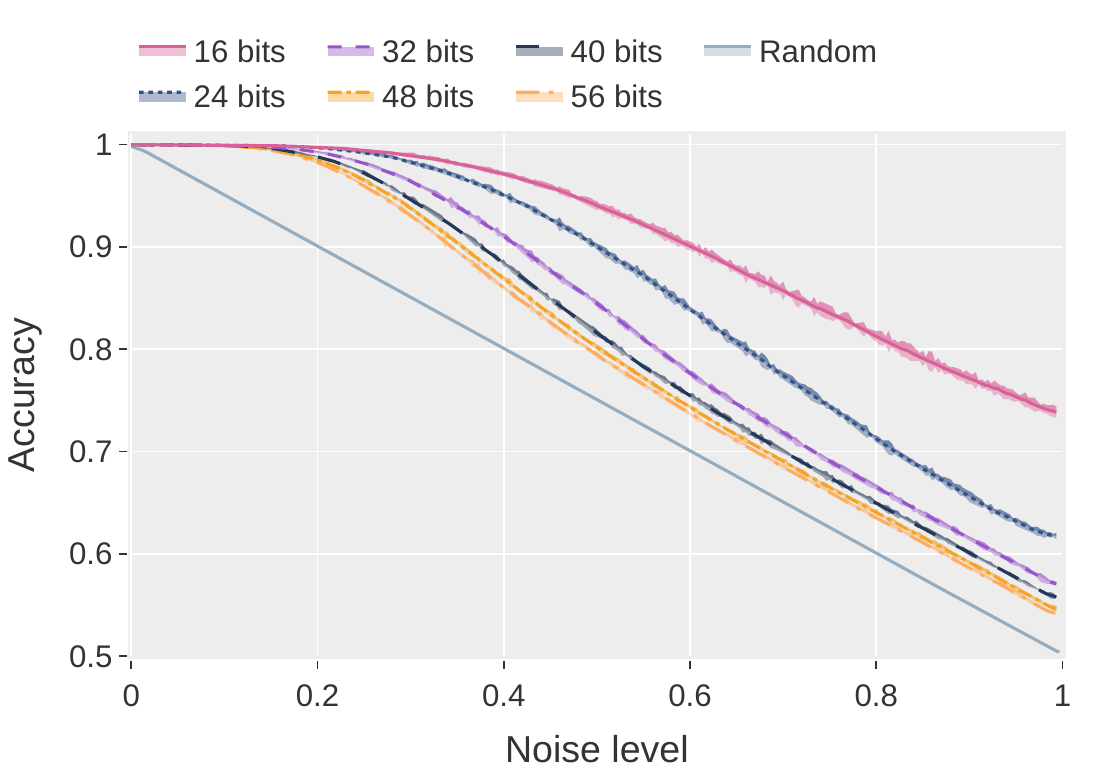}
        \caption{Bit prediction accuracy of the trained model depending on the noise level. The diagonal is a trivial baseline achieved by rounding.}
        \label{fig:prediction_acc}
    \end{minipage}
    \hfill
    \begin{minipage}[t]{.48\textwidth}
        \centering
        \includegraphics[width=1\columnwidth]{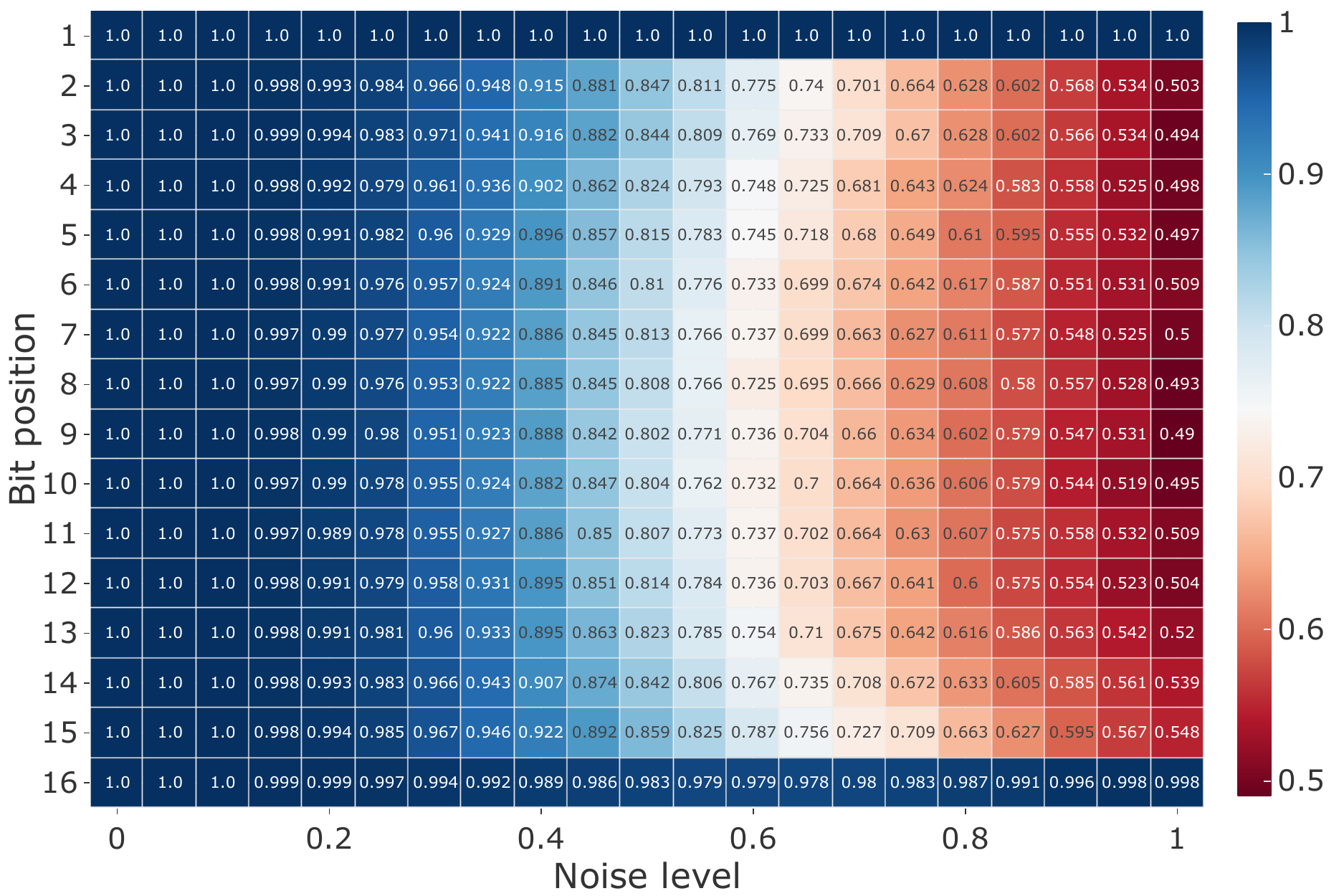}
        \caption{Prediction accuracy of each bit depending on the noise level for 16-bit factors of 32-bit numbers. We can see that for small noise all bits are predicted precisely, for large noise only the lowest and the highest bit can be predicted. }
        \label{fig:prediction_bits}
    \end{minipage}
\end{figure*}

We wish to apply the diffusion process to produce two integer numbers $a$,$b$ given their product $ab$. Such function cannot be directly approximated by a neural network for two reasons. First, this function is generally accepted to be extremely hard for which no efficient algorithm is known \cite{lenstra2000integer} and also our experiments confirm the inability of direct learning (see Fig.~\ref{fig:prediction_acc} portion with the large noise level, for evidence). Second, the factors are not unique prohibiting a straightforward supervised learning approach. 
Therefore we took the denoising approach where the neural network is asked to correct errors in a partially correct solution instead of outputting a fully correct solution from scratch. Such function we found to be learnable and also it allows obtaining samples from the whole distribution of factors, not only one particular. 

To make the diffusion process work well for factorization, several modifications to the standard diffusion schema are introduced. At first, all functions need to be conditioned on the given number $ab$ to be factored. We augment the neural network with an additional input in which $ab$ is represented in binary one-hot encoding. Similarly, each $x_t$ consists of one-hot encoded $a$ and $b$: $\hat{a},\hat{b} = \hat{x}_0=\mu(x_t, \bar{\alpha}_t, ab)$. 

To train the model, we generate training examples consisting of two odd random numbers of $n/2$ bits each (by selecting each bit randomly, hence some leading bits may be zeros), calculate their product $ab$ and form $x_0$ by concatenating one-hot encodings of $a$ and $b$. Then, we sample uniformly $\bar{\alpha}_t$ and obtain $x_t$ by sampling $q(x_t|x_0)$ as given in Eq.~\ref{eq:qsample}. All three inputs to the neural network are concatenated along the feature axis forming a sequence of length $n$ (the inputs $ab$ and $x_t$ are $n$ bits long and $\bar{\alpha}_t$ is replicated in each sequence position). The neural network $\mu(x_t, \bar{\alpha}_t, ab)$ is trained using KL divergence given by Eq.~\ref{eq:KL} to estimate $x_0$. We use a dataset of 10M examples generated this way, smaller datasets in our setup lead to overfitting. 

When the model is trained, we can use it for sampling to find the factors of a given number $ab$. For testing, we use composite numbers having exactly two prime factors each of length roughly $n/2$. We form a test set of 1K examples and explicitly make sure that none of these factors are used as multiplicands in the training set. 

Diffusion models are meant for producing samples of some distribution. Here, the distribution, conditioned on the number $ab$ to be factorized, consists of all the factorizations of $ab$. During training, we use randomly generated $a$ and $b$ which may be composite numbers themselves, so the neural network learns to factorize any composite number. But we test only on the most interesting (and possibly the hardest) case when both $a$ and $b$ are primes. So the distribution to sample from at test time consists of two discrete points $(a, b)$ and $(b,a)$.

The sampling algorithm is inspired by the reverse diffusion but it has two differences:  (a) it uses additive update instead of multiplicative and (b) it retains the full probability distribution from step to step and performs sampling from it only to present data to the neural network.
Given the total number of steps $T$, the algorithm (see Algorithm~\ref{alg:sampling}) works backward from step $T$ toward step 1. The initial distribution $x_T$ is created with equal probabilities of 0 and 1 for each bit and then at each step a sample $x_{sample}$ from it is drawn which is presented to the neural network $\mu$. The neural network returns an approximation of the correct bits $\hat x_0$. The probabilities of the previous timestep $x_{t-1}$ are calculated by taking the weighted average of $x_{t}$ and $q(x_{t-1}|\hat x_0)$ with a coefficient $\gamma$. We use $\gamma=0.9$. The algorithm returns the bit probabilities at the final step (or at an intermediate step if the solution is found) which should be close to binary if the neural network produces a good approximation, but in practice, we take the argmax over its category dimension as the bit values. This algorithm assumes that the neural network $\mu$ returns probabilities that sum to 1 for each bit; that is achieved by placing softmax as the last layer in $\mu$. 

The motivation behind deviating from the standard sampling algorithm is that the neural network produces a very inexact approximation $\hat{x}_0$ which by Eq.~\ref{eq:3} is often converging to a pair of numbers $a$, $b$ whose product is not the required $ab$. Also, the multiplicative nature of Eq.~\ref{eq:3} prohibits recovering from confident errors i.e. if, at some step the network gives a wrong 0 in some position, it is almost impossible to get it to 1 by using multiplication in the subsequent steps. The proposed modifications remedy these pitfalls. Use of the additive update allows easy recovery from confident mistakes and keeping the full probabilities instead of one sample retains more information and places less weight on each individual (and possibly wrong) update.   
A drawback of these modifications is that we lose the diversity of samples. That means that we may get only one factorization solution of $ab$ instead of all of them. But we are happy with that since there is essentially only one way to factor examples in the test set. 
We have confirmed experimentally that the proposed sampling algorithm works better.
%, see Fig.~\ref{fig:sampling_ablation} in the Appendix~\ref{sec:ablation}. 

\begin{algorithm}[h!]
\small
\caption{Sampling}

\begin{algorithmic}[1]
\State $x_T = 0.5^{n\times2}$  
%\Comment{Initialize $x_T$ with equal probabilities for each bit}
%\item[]
\For{$t=T$...1}
    \State $\bar{\alpha}_{t-1} = 1-(t-1)/T$
    \State $x_{sample} \sim \mathcal{C}(x_t)$
    \State $\hat{x}_0= \mu(x_{sample}, \bar{\alpha}_{t-1}, ab)$
     \If{$a$ and $b$ encoded in $\hat x_0$ multiply to $ab$}
            \Return $\hat x_0$
        \EndIf
    \State $x_{t-1} = \gamma x_t +(1-\gamma)[\bar{\alpha}_{t-1} \hat{x}_0 + (1-\bar{\alpha}_{t-1})/K]$
\EndFor
\State \Return $x_0$
\end{algorithmic}
\label{alg:sampling}
\end{algorithm}

Another modification is that we use relaxed categorical distribution (with a temperature equal to 1) instead of categorical. Their difference is how samples are produced. Categorical distribution introduces discrete noise characterized as bit flips. Relaxed distribution has more fine-grained noise which facilitates training and obtains a better success rate at inference. To work with the relaxed distribution we employ all the formulas given above except for sample generation where we apply the Gumbel Softmax technique \cite{jang2016categorical}. 
We have confirmed experimentally that using the relaxed distribution is beneficial.
%, see  Fig.~\ref{fig:discrete_vs_relaxed} in the Appendix~\ref{sec:ablation}.  

\section{Choice of the Neural Model}
The architecture of the neural network used to perform denoising has a significant impact on the overall performance. As the integer numbers are encoded as binary sequences, the task is of sequence-to-sequence nature. We evaluated three suitable architectures: Transformer \cite{vaswani2017attention}, Neural GPU \cite{Kaiser2015NeuralGPU} and Residual Shuffle-Exchange Networks (RSE) \cite{shuffle-exchange,draguns2021residual} but were not satisfied with their performance on our task. Therefore we tried to implement a new architecture, which turned out to be about 10\% better than Transformer and RSE.
 % see Fig.\ref{fig:architectures} in the Appendix~\ref{sec:ablation}. 

The new architecture is a sequence-to-sequence recurrent convolutional neural network based on the combination of Neural GPU and Shuffle-Exchange networks. It operates on hidden state $s\in \mathbb{R}^{n \times m}$ where $n$ is the sequence length and $m$ is the number of feature maps. It applies the following transformation, named Convolutional Shuffle Unit (CSU), to the sequence transforming the state $s_r$ at the recurrent-step $r$ to the state at the next step $s_{r+1}$:

\centerline{$\begin{aligned}
s_{drop} &= dropout(s_r) \\
s_F &= \text{ForwardShuffle}(s_{drop}) \\
s_R &= \text{ReverseShuffle}(s_{drop}) \\
g &= \text{GELU}(\text{InstanceNorm}(W\circledast[s_{drop} | s_F | s_R]+B))\\
c &= W' g + B' \\
s_{r+1} &= \sigma(S)\odot s_r + Z \odot c \\
\end{aligned}$}

In the above equations, $W$ is a convolution weight matrix of size $3m \times h \times 3$, where $h$ is the hidden size; $W'$ is a linear transformation weight matrix of size $h \times m$; $S$ and $Z$ are vectors of size $m$; $B$ and $B'$ are biases $-$ all of those are learnable parameters; $\odot$ denotes element-wise vector multiplication and $\sigma$ is the sigmoid function, $\circledast$ denotes convolution, | denotes concatenation along the feature axis. We choose the hidden size $h=4m$. 

The CSU starts by regularizing its input with dropout \cite{srivastava2014dropout} (we use dropout rate 0.1), then the input is concatenated with its forward- and reverse-shuffled versions \cite{shuffle-exchange} (ForwardShuffle and ReverseShuffle, accordingly). The forward shuffle divides the sequence into halves and interleaves the halves. The reverse shuffle does the opposite -- places even elements consecutively in the first half of the sequence and the odd elements in the second half. 
Convolution of kernel size 3 is then applied to the concatenated sequence followed by Instance Normalization \cite{ulyanov2016instance} and GELU \cite{hendrycks2016gaussian}. It is shown in \cite{shuffle-exchange} that repeated application of forward or reverse shuffle together with combining adjacent sequence elements allows rapid (in $O(\log n)$ steps) flow of information between any, possibly distant, sequence positions. In the new model, we use both shuffles together to even more facilitate long-range information flow and combine them with convolution to deal with short-range interactions. 

The last step of the CSU is a scaled residual connection. The candidate $c$ is scaled by a zero-initialized parameter $Z$ and added to the scaled input state $s_r$. The input scale parameter $S$ is chosen such that $\sigma(S) \approx 0.95$. Scaling both values in this way was shown to allow stable training of very deep residual networks \cite{bachlechner2020rezero} and lead to excellent performance in recurrent networks \cite{zakovskis2021gates}. 

The whole neural architecture consists of the input projection part, the recurrent part, and the output projection part. In the input part, two linear layers with GELU in-between are applied to data in each input sequence position independently to obtain the initial hidden state $s_0$.  Then, CSU is applied $\max(n/2, 4\lceil\log_2(n)\rceil)$ times in a recurrent fashion sharing the same parameters. Such recurrent depth was chosen as a reasonable compromise between the high expressive power of a deep network and the faster training of a shallow one. Each position of the last state is projected by a linear mapping to two values followed by softmax to obtain the bit probabilities of $a$ and $b$.  

\section{Results}\label{sec:results}

We have implemented the proposed method in Tensorflow, the code is available at \url{https://github.com/KarlisFre/diffusion-factorization}. To present our main results, we trained the model to factor 16--56-bit numbers. The length 56 was chosen roughly to be the maximum for which we can get a non-zero success rate. The model itself is independent of the sequence length, so we create batches of inputs in length increments of 8 in the given range and one training step consists of minimizing the loss for all these batches simultaneously like in \cite{freivalds2017improving}. The training was performed for 1M steps using AdaBelief optimizer \cite{zhuang2020adabelief} taking 2 weeks on two NVIDIA RTX A6000 GPUs. We chose the number of feature maps $m=384$ yielding a model with 5.8M trainable parameters. This choice was obtained experimentally in a few trials to get good results within the given training budget. 

All the results presented below are the averages over a batch of 256 examples. We repeated each experiment 5 times with different batches and depict the standard deviation as a shaded area in the line charts. 
% For heatmaps, we give standard deviation numerically and include them in Appendix~1. 

At first, let us explore how well the neural model is able to learn denoising. Fig~\ref{fig:prediction_acc} shows bit accuracy depending on the amount of introduced noise which is equal to $1-\bar{\alpha}_t$. For small amounts of noise, we get good prediction accuracy. For large noise, the model performs only slightly better than the trivial baseline that rounds each bit of the input $x_t$. Poor prediction in case of large noise is expected to some extent because there can be two possible results $a$, $b$ or $b$, $a$, and with purely random $x_t$ it is impossible to determine which order is the right one.

We can inspect more closely which bits are well-predicted in Fig~\ref{fig:prediction_bits}. The figure shows the accuracy for each 16-bit factor of 32-bit numbers. The first bit is always correct since we use only odd numbers in training and testing. The last bit is also predicted accurately which is possible solely from the magnitude of the composite number $ab$. And there is a general tendency that the leading or trailing bits are better predicted than the middle ones. For larger numbers, the findings are similar, only the overall accuracy is lower.

\begin{figure*}[ht]
    \centering
    \begin{subfigure}[t]{.48\textwidth}
        \centering
        \includegraphics[width=1\columnwidth]{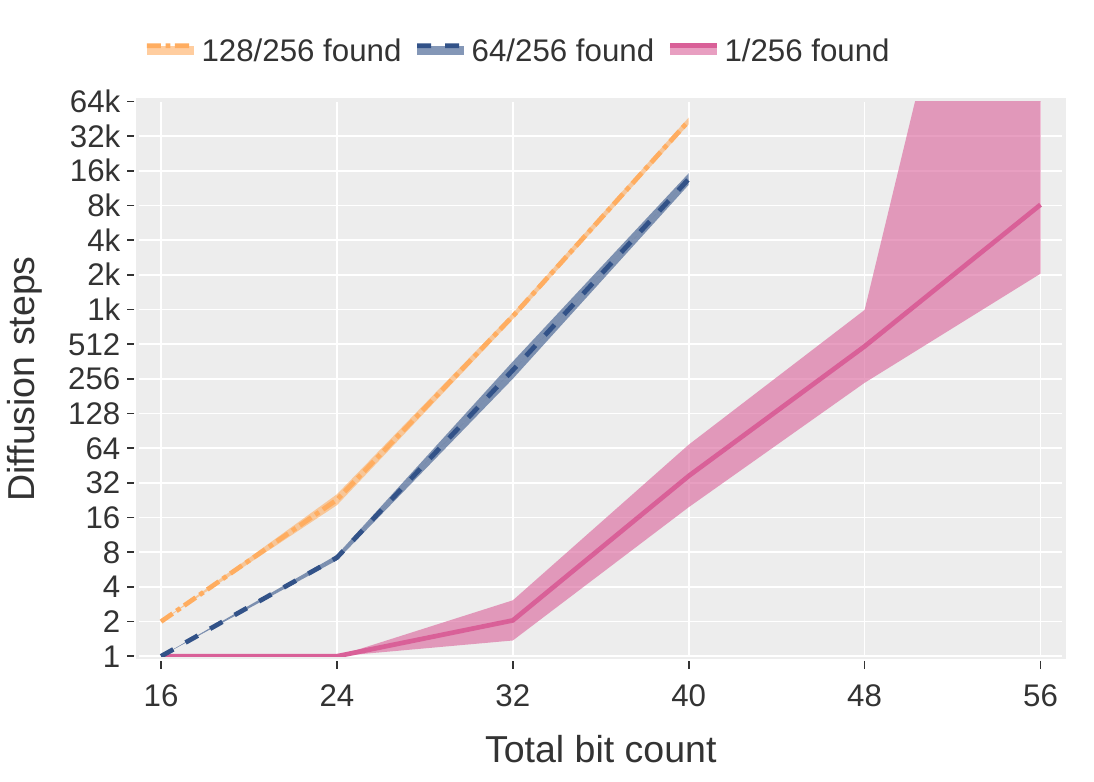}
        \caption{}
        % \caption{Diffusion steps taken to reach a given fraction of fully solved instances. Note that diffusion steps are presented log-scale. }
        \label{fig:length_vs_steps}
    \end{subfigure}
    \hfill
    \begin{subfigure}[t]{.48\textwidth}
        \centering
        \includegraphics[width=1\columnwidth]{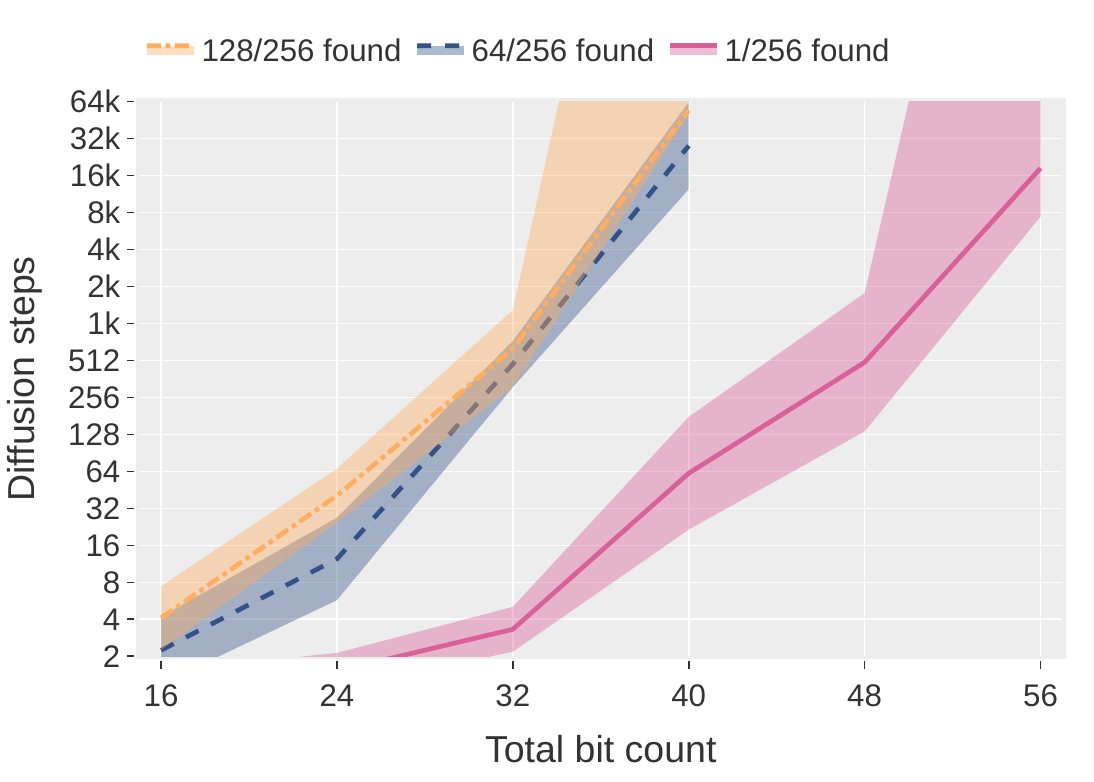}
        \caption{}
        % \caption{Diffusion steps taken to reach a given fraction of fully solved instances on batch consisting of same instance. Note that diffusion steps are presented log-scale.}
        \label{fig:length_vs_steps1}
    \end{subfigure}
    \caption{Diffusion steps taken to reach a given fraction of fully solved instances on batches containing different numbers (a) and on batches of equal numbers (b). Note that diffusion steps are presented log-scale.}
\end{figure*}

Next, we investigate the factorization performance depending on the bit-length of the numbers. Fig.~\ref{fig:length_vs_steps} shows how many diffusion steps are needed to factor a given fraction of examples in the batch. Since diffusion is a random process that may skip off an already found solution, we mark the example as factored if it happened at least in one of the diffusion steps. We see that small examples can be factored in a few steps but, for longer examples, the increase is exponential. Note that the model has indeed learned how to factor unseen numbers since we explicitly made sure that none of the prime factors used for testing were shown to the model during training. 

Often the goal is to factor one given number. We can replicate this number to fill the whole batch, process the batch in parallel, and expect that some of the replicas will be factorized faster due to randomness in sampling. Fig.~\ref{fig:length_vs_steps1} shows such a scenario. We see that the general shape of the lines is similar to those in Fig.~\ref{fig:length_vs_steps} only the variance is higher. The same mean in both these charts is expected since the data is the same, only replicated in the latter case. The variance is higher because some numbers appear to be easier to be factorized. We see that the line regarding $1/256$ is significantly below the two other lines even taking the variance into account, indicating that some instances (of the same number) in the batch get solved faster than the others, hence it is indeed useful to use such replicated batches. Also, at least one of the replicated numbers up to 48 bits got solved within the step limit showing that there are no hard numbers that the method is unable to factorize at all. 

%The good news is that all the tested numbers up to 40 bits eventually got factorized. 

\begin{figure*}[ht]
    \centering
    \includegraphics[width=0.48\textwidth]{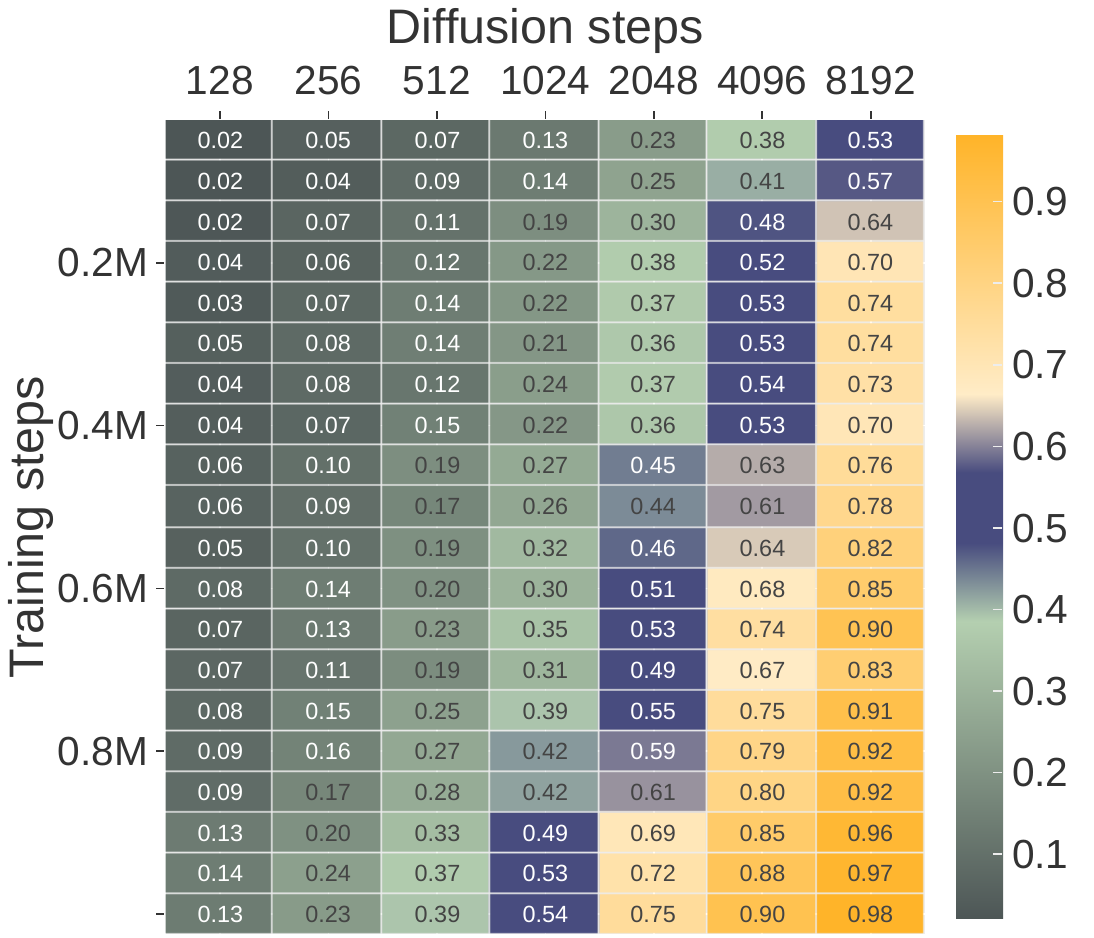}
    %\hspace{1cm}
    \hfill
    \includegraphics[width=0.48\textwidth]{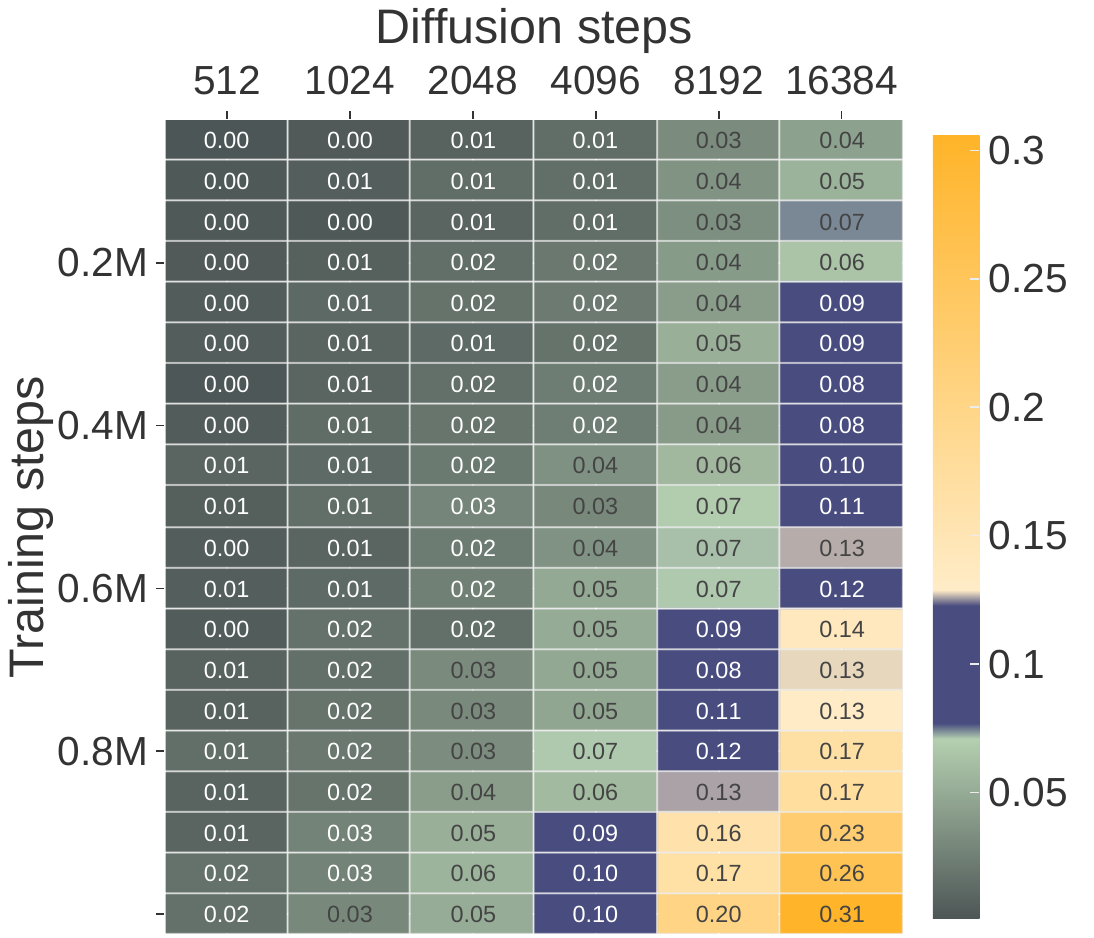}
    \caption{Success rate depending on training steps and diffusion steps on 32 bit (left) and 40 bit (right) integers. Note that diffusion steps are depicted in log-scale.}
    \label{fig:success}
\end{figure*}

It is interesting to analyze how much resources it is advisable to invest in training. If we invest more time in training, fewer diffusion steps (and computation time, respectively) are necessary to factor the given numbers. Notably, the one-time investment for training pays off for each number we wish to factor in afterward. Fig~\ref{fig:success} shows a heat map depicting the success rate (fraction of solved examples in the batch) depending on the training and diffusion steps for 32 and 40-bit numbers. We can see that the success rate increases both with training time and diffusion steps where the increase with training time is roughly linear. It can be observed that borders of equal success rate (one of such is marked in blue color) form almost straight lines. Since training steps are presented linearly but diffusion steps logarithmically, it means that a linear increase in training time leads to an exponential reduction of diffusion steps. This is very good news showing that investing in training pays off. The bad news is that the success rate decreases exponentially with sequence length.  

\section{Conclusion and Outlook} \label{conclusion}
We have presented an approach that solves the integer factorization problem using neural networks and discrete denoising diffusion.  We found that neural networks can learn error correction in integer factors and, although being too imprecise to be used directly, it can gradually arrive at the correct solution when applied iteratively, like in denoising diffusion. 
As a subsidiary result, we have presented a novel neural architecture that performs well on the denoising task. This new architecture, when properly validated, may find applications in other sequence-to-sequence tasks. 

We have analyzed how the method scales with respect to training and sampling steps and the bit-length of the numbers. Increasing the bit-length requires an exponential increase of the sampling steps at the inference or exponential investment in training to reach the same success rate suggesting that the factorization problem cannot be solved in polynomial time. On the other hand, investment in training gives exponential benefit during inference. So we might hint that a huge one-time investment in training would allow factorizing long numbers quickly afterward. The current research is limited to numbers up to 56 bits, a neural model with 5.8M trainable parameters, and 2 week training time on 2 GPUs. This is a very limited setup to fully understand the scaling behavior and proper relation between the two mentioned trends. The small integers considered in this paper can be factorized in a fraction of a second using standard methods while the neural approach can take several minutes (depending on the number of diffusion steps). So, further work with much more investment in computing resources is definitely needed to see whether the neural methods can present an asymptotic speedup. 

Although tested only on primes, the algorithm itself deals with a more general problem -- factoring the given integer into two multiplicands which themselves can be composite numbers. But virtually all classical factorization algorithms exploit in an essential way the additional information that the factors themselves are primes. A further direction for improvement could be investigating how to incorporate the properties of primality in the algorithm. 

The same denoising idea may be applied to other discrete search problems, for example, SAT. Currently, the main tool for solving them is tree-search. Diffusion, as employed here, is essentially a linear goal-directed randomized search that may serve as an alternative to the tree-search. We look forward to new results in this direction.

Considering a broader scope, an important theoretical question is whether a polynomial-time algorithm exists for factorization. Humans have not found such yet but, if it exists, could it be discovered automatically via learning a neural network? On the negative side, if a method someday would allow factoring long integers quickly, it will yield many cryptosystems insecure and secrets, currently protected by them, revealed. 

% \begingroup
% \let\clearpage\relax
% \bibliography{bibliography.bib}
% \endgroup

\bibliography{bibliography.bib}

\begin{thebibliography}{10}
\providecommand{\url}[1]{\texttt{#1}}
\providecommand{\urlprefix}{URL }
\providecommand{\doi}[1]{https://doi.org/#1}

\bibitem{austin2021structured}
Austin, J., Johnson, D., Ho, J., Tarlow, D., van~den Berg, R.: Structured denoising diffusion models in discrete state-spaces. Advances in Neural Information Processing Systems  \textbf{34} (2021)

\bibitem{bachlechner2020rezero}
Bachlechner, T., Majumder, B.P., Mao, H.H., Cottrell, G.W., McAuley, J.: {Rezero is all you need: Fast convergence at large depth}. arXiv preprint arXiv:2003.04887  (2020)

\bibitem{buhler1993factoring}
Buhler, J.P., Lenstra, H.W., Pomerance, C.: Factoring integers with the number field sieve. In: The development of the number field sieve, pp. 50--94. Springer (1993)

\bibitem{draguns2021residual}
Draguns, A., Ozoli{\c{n}}{\v{s}}, E., {\v{S}}ostaks, A., Apinis, M., Freivalds, K.: {Residual Shuffle-Exchange Networks for Fast Processing of Long Sequences}. In: Proceedings of the AAAI Conference on Artificial Intelligence. vol.~35, pp. 7245--7253 (2021)

\bibitem{freivalds2017improving}
Freivalds, K., Liepins, R.: Improving the neural {GPU} architecture for algorithm learning. The ICML workshop Neural Abstract Machines \& Program Induction v2 (NAMPI 2018)  (2018)

\bibitem{shuffle-exchange}
Freivalds, K., Ozoli\c{n}\v{s}, E., \v{S}ostaks, A.: {Neural Shuffle-Exchange Networks -- Sequence Processing in {O}($n$ log $n$) Time}. In: Advances in Neural Information Processing Systems 32, pp. 6626--6637. Curran Associates, Inc. (2019)

\bibitem{gaile2023unsupervised}
Gaile, E., Draguns, A., Ozoli{\c{n}}{\v{s}}, E., Freivalds, K.: Unsupervised training for neural tsp solver. In: Learning and Intelligent Optimization: 16th International Conference, LION 16, Milos Island, Greece, June 5--10, 2022, Revised Selected Papers. pp. 334--346. Springer (2023)

\bibitem{hendrycks2016gaussian}
Hendrycks, D., Gimpel, K.: {Gaussian Error Linear Units (GELUs)}. arXiv preprint arXiv:1606.08415  (2016)

\bibitem{ho2022cascaded}
Ho, J., Saharia, C., Chan, W., Fleet, D.J., Norouzi, M., Salimans, T.: Cascaded diffusion models for high fidelity image generation. Journal of Machine Learning Research  \textbf{23}(47),  1--33 (2022)

\bibitem{hoogeboom2021argmax}
Hoogeboom, E., Nielsen, D., Jaini, P., Forr{\'e}, P., Welling, M.: Argmax flows and multinomial diffusion: Learning categorical distributions. Advances in Neural Information Processing Systems  \textbf{34} (2021)

\bibitem{jang2016categorical}
Jang, E., Gu, S., Poole, B.: Categorical reparameterization with gumbel-softmax. arXiv preprint arXiv:1611.01144  (2016)

\bibitem{jansen2005neural}
Jansen, B., Nakayama, K.: Neural networks following a binary approach applied to the integer prime-factorization problem. In: Proceedings. 2005 IEEE International Joint Conference on Neural Networks, 2005. vol.~4, pp. 2577--2582. IEEE (2005)

\bibitem{Kaiser2015NeuralGPU}
Kaiser, {\L}., Sutskever, I.: Neural {GPU}s learn algorithms. arXiv preprint arXiv:1511.08228  (2015)

\bibitem{Diffwave}
Kong, Z., Ping, W., Huang, J., Zhao, K., Catanzaro, B.: {DiffWave: {A} Versatile Diffusion Model for Audio Synthesis}. In: 9th International Conference on Learning Representations, {ICLR} 2021, Virtual Event, Austria, May 3-7, 2021. OpenReview.net (2021), \url{https://openreview.net/forum?id=a-xFK8Ymz5J}

\bibitem{lenstra2000integer}
Lenstra, A.K.: Integer factoring. Towards a Quarter-Century of Public Key Cryptography: A Special Issue of DESIGNS, CODES AND CRYPTOGRAPHY An International Journal. Volume 19, No. 2/3 (2000) pp. 31--58 (2000)

\bibitem{meletiou2002first}
Meletiou, G., Tasoulis, D., Vrahatis, M.N., et~al.: {A first study of the neural network approach to the RSA cryptosystem}. In: IASTED 2002 Conference on Artificial Intelligence. pp. 483--488 (2002)

\bibitem{ozolins2022goal}
Ozolins, E., Freivalds, K., Draguns, A., Gaile, E., Zakovskis, R., Kozlovics, S.: Goal-aware neural sat solver. In: 2022 International Joint Conference on Neural Networks (IJCNN). pp.~1--8. IEEE (2022)

\bibitem{pollard1978monte}
Pollard, J.M.: Monte carlo methods for index computation. Mathematics of computation  \textbf{32}(143),  918--924 (1978)

\bibitem{rombach2021high}
Rombach, R., Blattmann, A., Lorenz, D., Esser, P., Ommer, B.: {High-Resolution Image Synthesis with Latent Diffusion Models}. arXiv preprint arXiv:2112.10752  (2021)

\bibitem{Shor1994AlgorithmsFQ}
Shor, P.W.: Algorithms for quantum computation: discrete logarithms and factoring. Proceedings 35th Annual Symposium on Foundations of Computer Science pp. 124--134 (1994)

\bibitem{silver2018general}
Silver, D., Hubert, T., Schrittwieser, J., Antonoglou, I., Lai, M., Guez, A., Lanctot, M., Sifre, L., Kumaran, D., Graepel, T., et~al.: A general reinforcement learning algorithm that masters chess, shogi, and go through self-play. Science  \textbf{362}(6419),  1140--1144 (2018)

\bibitem{sohl2015deep}
Sohl-Dickstein, J., Weiss, E., Maheswaranathan, N., Ganguli, S.: Deep unsupervised learning using nonequilibrium thermodynamics. In: International Conference on Machine Learning. pp. 2256--2265. PMLR (2015)

\bibitem{srivastava2014dropout}
Srivastava, N., Hinton, G., Krizhevsky, A., Sutskever, I., Salakhutdinov, R.: Dropout: a simple way to prevent neural networks from overfitting. The journal of machine learning research  \textbf{15}(1),  1929--1958 (2014)

\bibitem{ulyanov2016instance}
Ulyanov, D., Vedaldi, A., Lempitsky, V.: Instance normalization: The missing ingredient for fast stylization. arXiv preprint arXiv:1607.08022  (2016)

\bibitem{vahdat2021score}
Vahdat, A., Kreis, K., Kautz, J.: Score-based generative modeling in latent space. Advances in Neural Information Processing Systems  \textbf{34} (2021)

\bibitem{vaswani2017attention}
Vaswani, A., Shazeer, N., Parmar, N., Uszkoreit, J., Jones, L., Gomez, A.N., Kaiser, {\L}., Polosukhin, I.: {Attention is All you Need}. In: Guyon, I., {Luxburg U.V. et al.} (eds.) Advances in Neural Information Processing Systems 30, pp. 5998--6008. Curran Associates, Inc. (2017)

\bibitem{tensorflowmodelgarden2020}
Yu, H., Chen, C., Du, X., Li, Y., Rashwan, A., Hou, L., Jin, P., Yang, F., Liu, F., Kim, J., , Li, J.: {TensorFlow Model Garden}. \url{https://github.com/tensorflow/models} (2020)

\bibitem{zakovskis2021gates}
Zakovskis, R., Draguns, A., Gaile, E., Ozolins, E., Freivalds, K.: {Gates are not what you need in RNNs}. arXiv preprint arXiv:2108.00527  (2021)

\bibitem{zhuang2020adabelief}
Zhuang, J., Tang, T., Ding, Y., Tatikonda, S.C., Dvornek, N., Papademetris, X., Duncan, J.: {Adabelief optimizer: Adapting stepsizes by the belief in observed gradients}. Advances in neural information processing systems  \textbf{33},  18795--18806 (2020)

\end{thebibliography}

%\clearpage
\appendix

\section{Overview of the Algorithm}
\begin{figure*}[ht]
    \centering
    \includegraphics[width=0.9\textwidth]{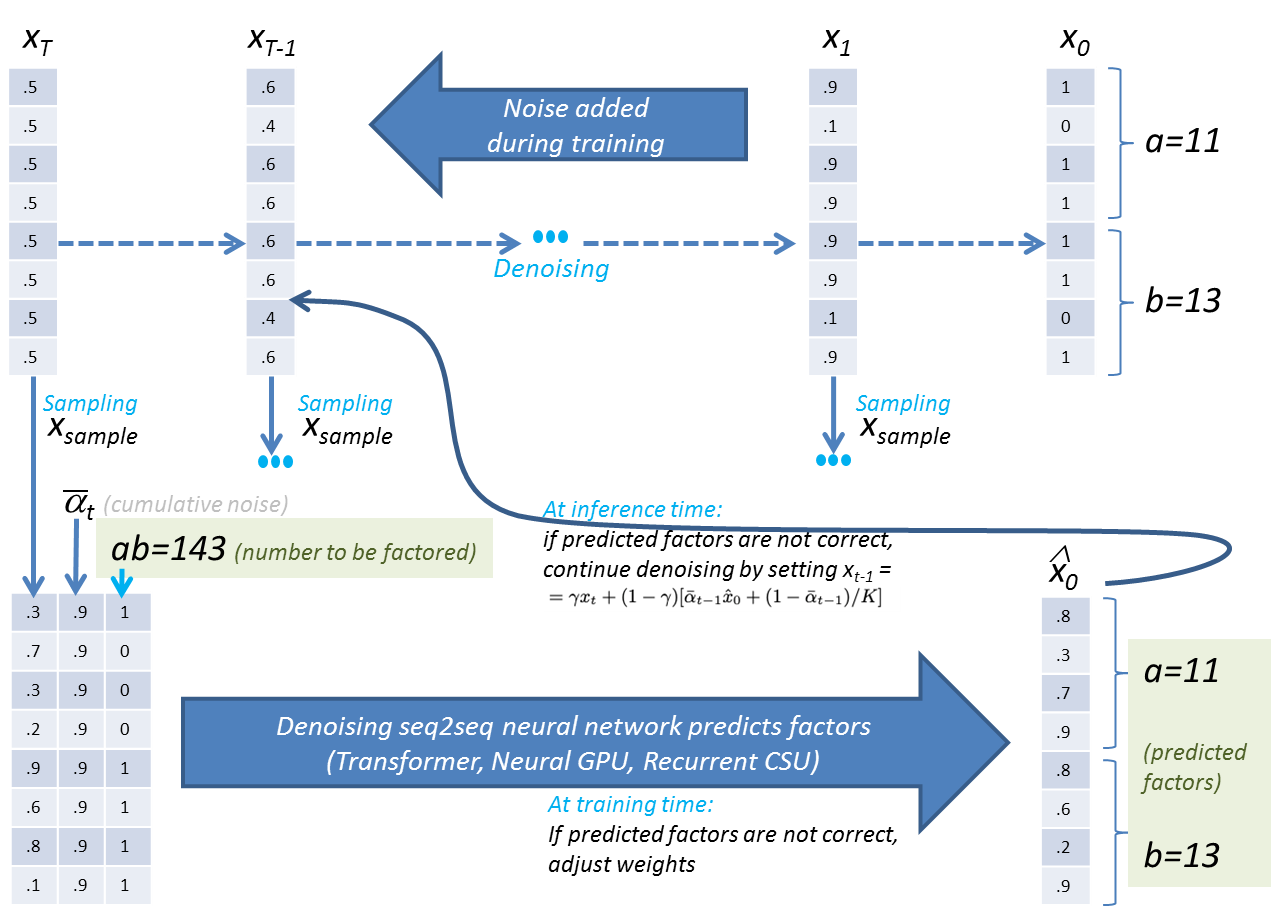}
    %\hspace{1cm}
    %\hfill
    %\includegraphics[width=0.48\textwidth]{figures/40_bit_mean.pdf}
    \caption{Illustration of the proposed diffusion approach for factorizing the number 143. A seq2seq neural network is trained to recover original vector $x_{0}$ (representing correct prime factors $a$ and $b$) from increasingly noised versions of this vector $x_{1}, x_{2},...,x_{T}$ given two additional inputs: the product of $a$ and $b$ ($ab$) and scalar $\bar{\alpha}_t$ characterising the cumulative noise level in the step $t$. The trained neural network later can be used to factor any number, e.g. $ab=143$ into its prime factors $a=11$ and $b=13$ by starting from the random vector $x_{T}$ and iteratively denoising it through repeated applying of the trained neural network given two additional inputs: product $ab=143$ and $\bar{\alpha}_t$. 
}
    \label{fig:overview}
\end{figure*}

\section{Failure Cases}\label{sec:failures}
In virtually all failure cases the algorithm converges to a pair of numbers whose product differs from the number to be factored only by a few bits. For a particular example, when factoring 3776028761 whose real factors are 59393 and 63577, the algorithm gives 59229 * 63789 that multiply to 3778158681. If we observe 3776028761 and 3778158681 in binary, they differ only in 4 bits.

3 776 028 761 = 0b11100001000100011010000001011001

3 778 158 681 = 0b11100001001100100010000001011001

Note that both returned multiplicands are composite numbers: 59229 = 3*3*6581 and 63789 = 3*11*1933. A  direction for improvement could be investigating how to incorporate the properties of primality in the algorithm.

\section{Design Choices and Ablations}\label{sec:ablation}
Here we validate our three main technical contributions: the new neural model, using relaxed distribution, and modifying the diffusion process -- that they are worthy, indeed. 

\subsection{Choice of the Model} \label{sec:model_ablation}
We have compared the proposed neural architecture with three suitable architectures -- Transformer \cite{vaswani2017attention}, Neural GPU \cite{Kaiser2015NeuralGPU} and Residual Shuffle-Exchange Networks (RSE) \cite{shuffle-exchange,draguns2021residual}. The Transformer is the most common architecture for sequence-to-sequence tasks and we use its standard implementation from TensorFlow Model Garden\cite{tensorflowmodelgarden2020}. The other two architectures have been designed especially for algorithmic tasks, showing good results for binary multiplication. We use the improved Neural GPU implementation by \cite{freivalds2017improving} and the official RSE implementation \cite{draguns2021residual}. 

\begin{figure*}[ht]
    \centering
    \begin{minipage}[t]{.32\textwidth}
        \centering
        \includegraphics[width=\columnwidth]{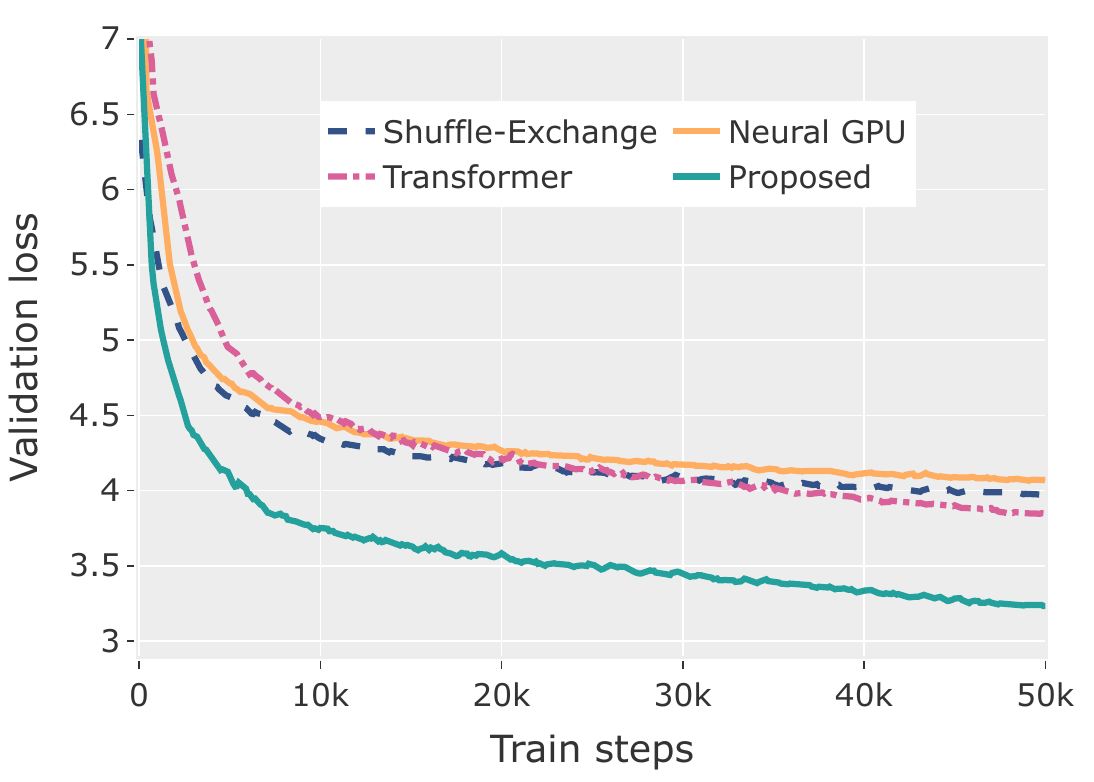}
        \caption{Validation loss during training for the compared architectures. The new architecture outperforms the others by a wide margin.}
        \label{fig:architectures}
    \end{minipage}
    \hfill
    \begin{minipage}[t]{.32\textwidth}
        \centering
        \includegraphics[width=\columnwidth]{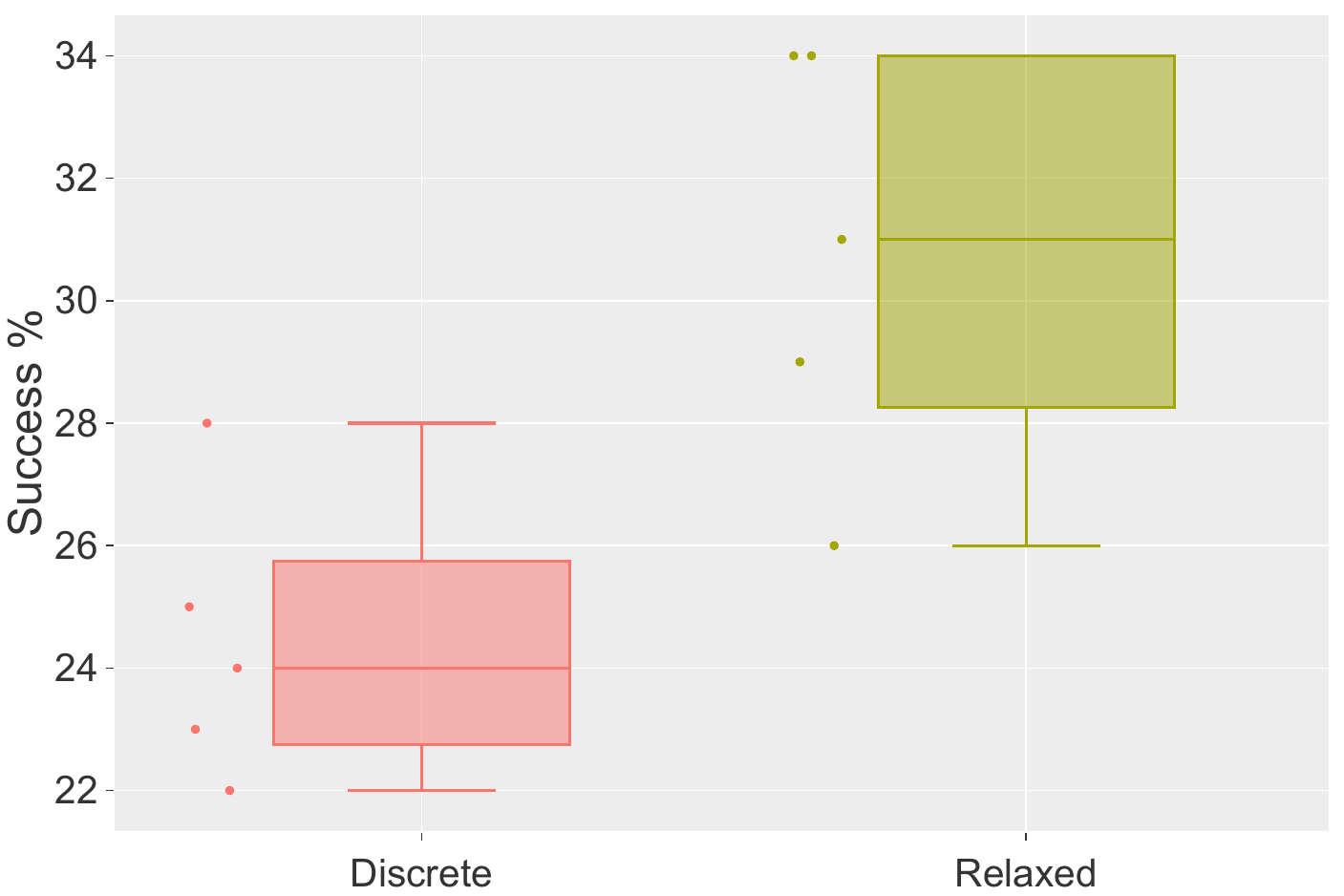}
        \caption{Box plot of success rate for discrete categorical vs. relaxed categorical distributions for factoring 32-bit numbers. The relaxed distribution performs better.}
        \label{fig:discrete_vs_relaxed}
    \end{minipage}
    \hfill
    \begin{minipage}[t]{.32\textwidth}
        \centering
        \includegraphics[width=\columnwidth]{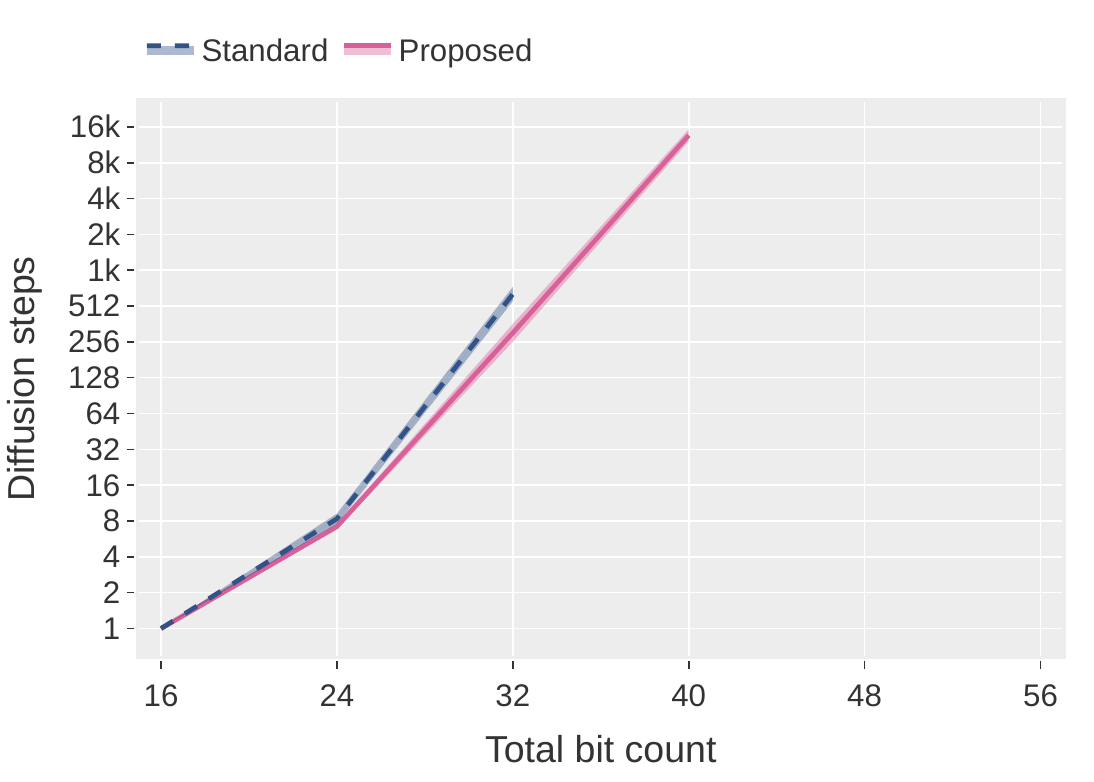}
        \caption{Comparing standard diffusion sampling and the proposed one. Diffusion steps (log-scale) taken to solve 25\% of the instances.}
        \label{fig:sampling_ablation}        
    \end{minipage}
\end{figure*}

We chose a smaller setup that can be trained in one day on one GPU. The numbers to factor were 32 bits long, training was done for 50K steps and the parameters of all the other models were chosen such that their inference speed roughly matches (also training speed was similar) to our model. The details are given in the Appendix~2.  The validation losses reached during training are depicted in Fig.\ref{fig:architectures}. We can see that the new architecture provides substantially lower loss. The loss reflects how well the model is able to correct the introduced noise and gives a good indication of its success later during sampling. 

Note that providing a new architecture should be considered as a subsidiary result of this paper. As we have done a fairly limited comparison without extensive hyperparameter tuning, we do not claim that the proposed one is the best possible for this task. 

\subsection{Relaxed vs. Discrete Distribution}
We use the same smaller setup as for comparing the neural architectures and trained our model with both (discrete) categorical and relaxed categorical distributions. Test results for 32-bit numbers having 5 trials for each of them is given in Fig.~\ref{fig:discrete_vs_relaxed}. From these box plots, we see that the relaxed distribution achieves a significantly higher success rate. 

\subsection{Modified Sampling}

We have compared the proposed sampling algorithm (Algorithm~\ref{alg:sampling}) with the standard one (as given in Sec.~\ref{sec:background}). We use the full model trained as described in Sec.~\ref{sec:results} and compare how many sampling steps are required with each of them to solve 25\% of the given instances. In Fig.~\ref{fig:sampling_ablation} we see that both sampling methods can do the task but the proposed sampling algorithm does that with significantly fewer steps. Importantly, the proposed sampling is able to factor 40-bit numbers within 64K steps, what is the limit for this figure, but the standard one cannot.

\end{document}